\documentclass[12pt]{article}
\usepackage[utf8]{inputenc}
\usepackage{xcolor}
\usepackage{float}
\usepackage[justification=centering]{caption}
\usepackage[margin=0.5in]{geometry}
\usepackage[pdftex]{graphicx}
\graphicspath{{\Documents}}     
\DeclareGraphicsExtensions{.pdf,.jpeg,.png}
\usepackage{titlesec}
\usepackage{moreverb}
\usepackage{makecell}
\usepackage[utf8]{inputenc}
\usepackage{multicol}
\usepackage{amsmath,amssymb,amsfonts}
\usepackage{algorithmic}
\usepackage{textcomp}
\usepackage{cite}
\usepackage{multirow}
\usepackage{hyperref}
\usepackage[T1]{fontenc}
\usepackage[utf8]{inputenc}

\newcommand{\mcT}{\mathcal{T}}

\title{\textbf{Equipment Health Assessment: Time Series Analysis for Wind Turbine Performance}}

\begin{multicols}{2}
\author{Jana Backhus\\
Industrial AI Lab\\
Hitachi America, Ltd., R\&D\\
Santa Clara, USA\\
\{Firstname.Lastname\}@hal.hitachi.com\\
\and
Aniruddha Rajendra Rao*\\
Industrial AI Lab\\
Hitachi America, Ltd., R\&D\\
Santa Clara, USA\\
Aniruddha.Rao@hal.hitachi.com\\
\and
Chandrasekar Venkatraman\\
Industrial AI Lab\\
Hitachi America, Ltd., R\&D\\
Santa Clara, USA\\
\vspace{1cm}
\{Firstname.Lastname\}@hal.hitachi.com\\
\and
Abhishek Padmanabhan\\
Atria University\\
India\\
abhishekp@atriauniversity.edu.in\\
\and
A.Vinoth Kumar\\
Atria Brindavan Power Private Limited\\
India\\
\vspace{1cm}
Vinoth.Kumar@atriapower.com\\
\and
Chetan Gupta\\
Industrial AI Lab\\
Hitachi America, Ltd., R\&D\\
Santa Clara, USA\\
\vspace{1cm}
\{Firstname.Lastname\}@hal.hitachi.com\\}
\end{multicols}

\date{}

\begin{document}

\maketitle
\vspace{-1.5cm}
\hspace{0.35cm}\textbf{Keywords:} SCADA Data, Time Series, Wind Turbine, Prediction, Classification, Ensemble.

\abstract{In this study, we leverage SCADA data from diverse wind turbines to predict power output, employing advanced time series methods, specifically Functional Neural Networks (FNN) and Long Short-Term Memory (LSTM) networks. A key innovation lies in the ensemble of FNN and LSTM models, capitalizing on their collective learning. This ensemble approach outperforms individual models, ensuring stable and accurate power output predictions. Additionally, machine learning techniques are applied to detect wind turbine performance deterioration, enabling proactive maintenance strategies and health assessment. Crucially, our analysis reveals the uniqueness of each wind turbine, necessitating tailored models for optimal predictions. These insight underscores the importance of providing automatized customization for different turbines to keep human modeling effort low. Importantly, the methodologies developed in this analysis are not limited to wind turbines; they can be extended to predict and optimize performance in various machinery, highlighting the versatility and applicability of our research across diverse industrial contexts.}

\section{Introduction}

In an era marked by mounting concerns about the environmental impact of traditional energy sources, the quest for sustainable and renewable alternatives has taken center stage in the global discourse. The imperative to reduce greenhouse gas emissions, secure energy independence, and mitigate the effects of climate change has driven a fundamental shift in our approach to power generation \cite{Gernaat2021ClimateCI, Koch2015TheIO}. At the forefront of this transformative journey stands wind energy, a clean and abundant resource harnessed to provide a solution to our growing energy needs while reducing our carbon footprint \cite{PortAgel2019WindTurbineAW, Pryor2020ClimateCI}. Wind turbines gracefully punctuate landscapes symbolizing our commitment to a cleaner and more sustainable future. However, the pursuit of harnessing the power of the wind comes with its own set of challenges \cite{Veers2019GrandCI, Yang2014WindTC, Narasinh2024InvestigatingPL}. The intermittent nature of wind, the need for effective grid integration, and the demands of operating and maintaining these complex machines have sparked a revolution in how we manage and optimize wind energy facilities.

At the heart of this operation and maintenance revolution lies the Supervisory Control and Data Acquisition (SCADA) system, a vital component that allows us to remotely monitor, control, and collect crucial data from wind turbines \cite{Jin2021ConditionMO, TautzWeinert2017UsingSD}. The insights offered by data obtained from SCADA systems ensure the efficient operation of wind turbines but also play a pivotal role in their longevity.  In this study, we leverage SCADA data from 13 wind turbines located in a wind farm in India to predict the power output, employing advanced time series methods, specifically Functional Neural Networks (FNN) \cite{Rao2021NonlinearFM, Rao2021ModernNF} and Long Short-Term Memory (LSTM) networks \cite{Hochreiter1997LongSM}. A key innovation lies in an ensemble of FNN and LSTM model to capitalize on their collective learning. This approach outperforms the individual models, ensuring high accuracy of the power output predictions when wind turbine is in a good state (good timeline) and significantly lower accuracy when the wind turbine is in a bad state (bad timeline). Then, statistical techniques are applied to the prediction performance errors to detect wind turbine deterioration, enabling proactive maintenance strategies and health assessment.
Our analysis leads to the important understanding that the uniqueness of each wind turbine necessitates tailored models for optimal predictions. This highlights the significance of offering automated customization for different turbines to minimize human efforts for modeling. Importantly, the methodologies devised in this analysis are not restricted to wind turbines; they can be extended to various other machinery to enhance their performance. This showcases the versatility and relevance of our research across a wide range of industrial settings.

This research is part of an effort to leverage equipment related data to analyze the health of the equipment where multiple entities are placed in a similar environment, but each entity is showing slightly different behavior \cite{Zhang2019DataDrivenMF, Zio2013PrognosticsAH}. The goal of this study is to develop methodologies that are not limited to our first target of wind turbines but can be extended to predict and optimize equipment performance in various machinery. \textbf{Furthermore, we assume that it will be too much human effort to fine-tune a prediction model for each machine entity and therefore we explore machine learning methodologies that are robust enough to perform considerably well over all entities. }In summary, this is an initial study that we hope to extend into a more versatile and applicable research across diverse industrial contexts in the future.

%%%%%%%%%%%%%%%%%%%%%%%%%%%%%%%%%%%%%%%%%%
\section{Data and Methods}

% \textcolor{red}{should we add source for this image below? --> The image is from Hitachi techshare..not sure if they created it themselves or if it was taken from somewhere. Maybe we should look into a replacement image and delete this one}

% \begin{figure}[H]
% \centering
% \includegraphics[width=8.5 cm]{Screenshot 2024-01-30 at 11.20.29 AM.png}
% \caption{Description of the Rotating Equipment Data\label{fig1}}
% \end{figure}  

% \textcolor{red}{add figure about SCADA for wind turbine?? OR just delete picture and don't add new one,  have to add new one if we wanna keep it, currently no link to figure}

\subsection{Data}
A SCADA dataset for wind turbine power generation collects real-time operational, environmental, anomalous, electrical, and communication data from individual turbines or across an entire wind farm \cite{MaldonadoCorrea2020UsingSD, Pandit2022SCADADF}. It encompasses turbine status (on/off, power output, rotor speed), environmental conditions (wind speed, direction, temperature), alarms for irregularities, electrical parameters (voltage, current), and communication details. The initial SCADA dataset measures approximately 2000 different features. We have limited our scope of interest, however, before downloading the data from a data collection server \cite{Narasinh2024InvestigatingPL}, to keep the data size small. Therefore, we have excluded all features that are metadata like timestamps or Boolean values related to wind turbine status.  For our study, we are interested in the power generation performance of each individual wind turbine. Therefore, we mainly focus on the measurements of actual power output and any related factors. We have chosen relevant parameters based on the following turbine components: e.g., rotor, pitch, gearbox, nacelle, power, and environmental information like wind \cite{Wang2022FaultDA}.

\begin{figure}[H]
\includegraphics[width=13.5 cm]{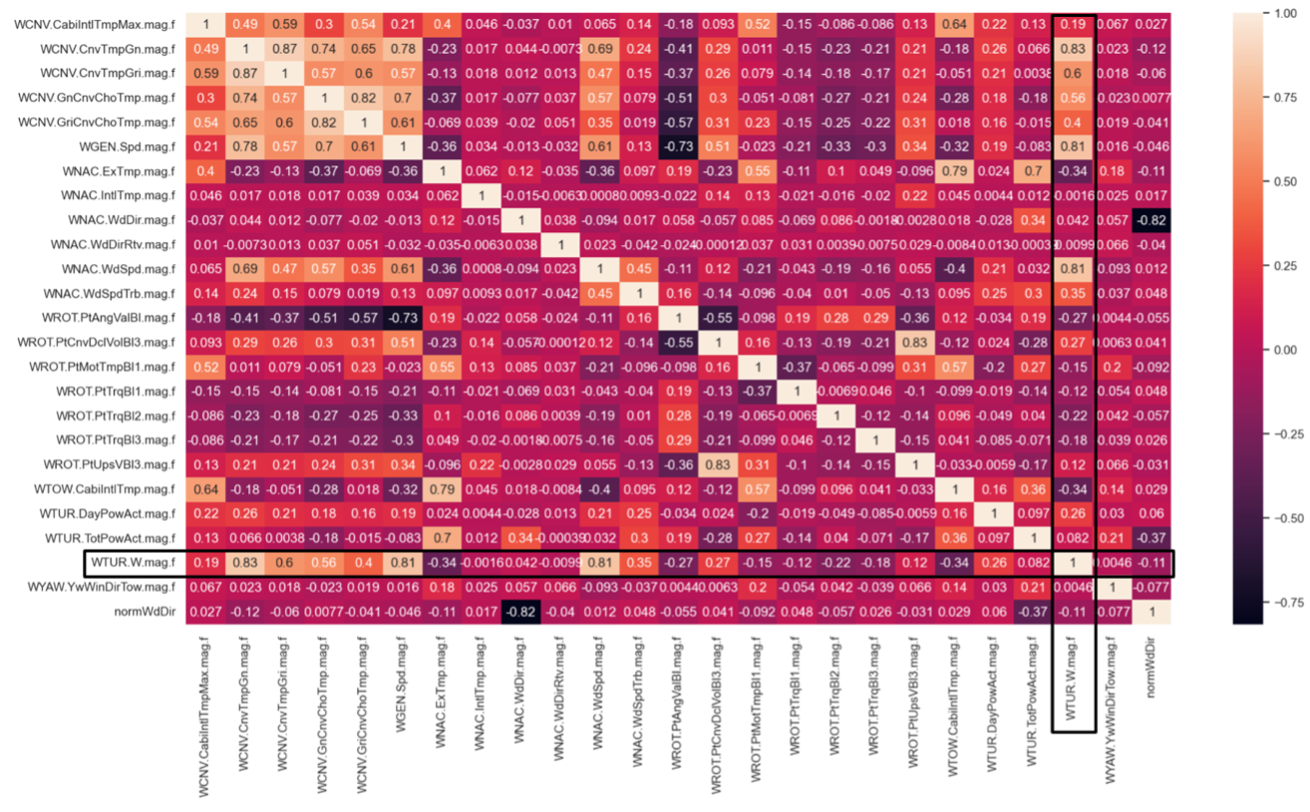}
\caption{Correlation Heatmap between Features of Interest}\label{fig2}
\end{figure}

After downloading an initial dataset with 76 features, we have conducted an exploratory data and correlation analysis on one wind turbine (no. 5) to further downsize the dataset of interest. The wind turbine was selected based on domain information that it has active pitch alignment and is considered one of the better performing wind turbines. We were able to decrease the feature set to 24 features, omitting 9 features because of rare changes in values and 43 features because of very high correlation (larger than 0.9) to at least one other feature. The correlation heatmap of the remaining features of interest is shown in Figure \ref{fig2}. Here, we highlighted the row and column for the actual power output (“WTUR.W.mag.f”) which is our target feature of interest to evaluate the power generation performance of a wind turbine. We also added an additional feature to the heatmap which is a calculation of the sinus radius of the wind direction, where the original wind direction values are given in degrees.

\begin{figure}[H]
\includegraphics[width=13.5 cm]{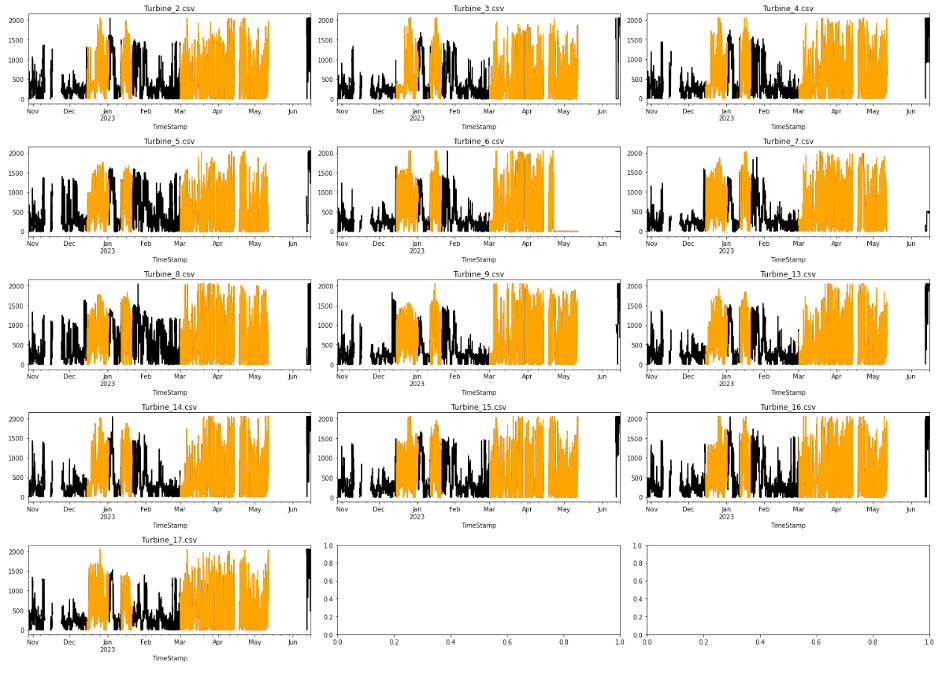}
\caption{Actual power output of the 13 wind turbines}\label{fig3}
\end{figure}

Figure \ref{fig3} shows the actual power output of the 13 wind turbines over the available time frame from end of October 2022 to early June 2023. The timelines colored in orange are considered “good” based on initial information received from the data provider. We also got informed that the whole wind farm is known to show "bad" performance between mid of November to mid of December 2022 due to dust collection. Furthermore, we can observe in Figure \ref{fig3} that there are several periods of missing data mostly caused by temporary shutdowns of the SCADA data collection system.

\subsection{Methods}

In predicting power outputs, traditional and statistical methods have long been employed to model relationships between variables (\cite{wang2021comparison}, \cite{dolara2015comparison}, \cite{theocharides2018machine}, \cite{massidda2017use}). Traditional machine learning methods like linear regression are fundamental tools, assuming a linear relationship between predictors and the power function output \cite{backhus2022cooling}. Polynomial regression, an extension of linear regression, can capture more complex relationships by introducing polynomial terms. These methods, while interpretable and computationally efficient, might struggle to capture intricate nonlinear patterns present in power functions, limiting their accuracy \cite{Parmezan2019EvaluationOS}. They also ignore the temporal nature of the data. We therefore consider two different time series prediction approaches using Deep Learning (DL) methods, which have been show to be most successful in the literature \cite{Wang2020ANF, Rao2021ModernNF, Song2020TimeseriesWP, Jin2019PredictionFT}, for our wind turbine power output prediction as described in the following two sections.

\subsubsection{Long Short-Term Memory (LSTM)}
Recently, deep learning methods, especially recurrent neural networks (RNNs) and long short-term memory networks (LSTMs) \cite{Hochreiter1997LongSM}, have revolutionized the time series prediction landscape \cite{Song2020TimeseriesWP, Lindemann2021ASO}. When applied to power function prediction, these DL methods can capture intricate dependencies and nonlinearities inherent in the data. Additionally, convolutional neural networks (CNNs) can be employed to capture spatial patterns within the data \cite{Jin2019PredictionFT}. Deep learning models, while computationally intensive, offer the advantage of automatic feature extraction, enabling them to discern complex relationships, and making them highly effective in accurately predicting power functions even in dynamic and intricate scenarios. Their ability to handle vast amounts of data and learn hierarchical representations often leads to superior performance in power function prediction tasks \cite{Han2019ARO}.

\begin{figure}[H]
\includegraphics[width=8.5 cm]{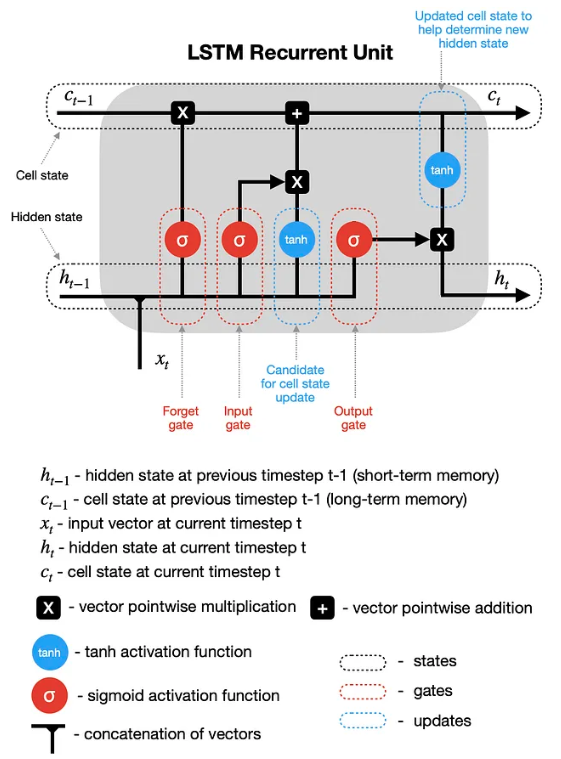}
\caption{LSTM Cell (\href{https://towardsdatascience.com/lstm-recurrent-neural-networks-how-to-teach-a-network-to-remember-the-past-55e54c2ff22e}{Source})}\label{fig4}
\end{figure}  

LSTM is a type of RNN architecture, specifically designed to capture long-term dependencies in sequential data, making them ideal for predicting power functions over time as seen in Figure \ref{fig4}. Traditional RNNs often struggle with learning patterns in data sequences that are separated by many time steps. LSTMs were introduced to address this issue by incorporating memory cells and sophisticated gating mechanisms.
LSTMs are widely used in various applications such as natural language processing, speech recognition, and time series prediction due to their ability to capture long-term dependencies in sequential data. LSTM components are as follows:\\
\begin{enumerate}
    \item Hidden state \& new inputs — hidden state from a previous timestep ($h_{t-1}$) and the input at a current timestep ($x_t$) are combined before passing copies of it through various gates.\\
    \item Forget gate — this gate controls what information should be forgotten. Since the sigmoid function ranges between 0 and 1, it sets which values in the cell state should be discarded (multiplied by 0), remembered (multiplied by 1), or partially remembered (multiplied by some value between 0 and 1).\\
    \item Input gate helps to identify important elements that need to be added to the cell state. Note that the results of the input gate get multiplied by the cell state candidate, with only the information deemed important by the input gate being added to the cell state.\\
    \item Update cell state —first, the previous cell state ($c_{t-1}$) gets multiplied by the results of the forget gate. Then we add new information from [input gate × cell state candidate] to get the latest cell state ($c_t$).\\
    \item Update hidden state — the last part is to update the hidden state. The latest cell state ($c_t$) is passed through the tanh activation function and multiplied by the results of the output gate.\\
    \item Finally, the latest cell state ($c_t$) and the hidden state ($h_t$) go back into the recurrent unit, and the process repeats at timestep t+1. The loop continues until we reach the end of the sequence. 

\end{enumerate}

\subsubsection{Functional Neural Networks (FNN)}
On the other hand, Functional data analysis (FDA) \cite{Ramsay1997FunctionalDA, Kokoszka2017IntroductionTF, Ferraty2011TheOH} is a branch of statistics that deals with data that are functions or curves, rather than simple numeric or categorical values. In FDA, the data is treated as a continuous function, and the goal is to analyze and model the behavior of the function over time or space. FDA methods allow researchers to extract meaningful information from functional data, such as trends, patterns, and underlying structures. This information can then be used to develop models and make predictions or forecasts.
A Functional Linear Model (FLM) \cite{Wang2016FunctionalDA, Reiss2017MethodsFS} is a statistical model used in FDA that extends the linear regression model to functional data. It is designed to handle data that consists of a collection of curves or functions, rather than discrete data points. The FLM assumes that the relationship between the response variable and the predictor variables can be represented by a linear function, but with a functional rather than scalar form. The predictor variables can be either functional or scalar and can be continuous or categorical. In FLM, the functional form of the predictor variable is expressed using a basis expansion of a finite set of basis functions. These basis functions can be any set of orthogonal functions that span the space of functions under consideration. For example, Fourier, wavelet, or B-spline basis functions can be used.

\begin{figure}[H]
\includegraphics[width=8.5 cm]{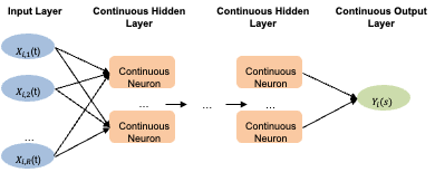}
\caption{The general architecture of Functional Neural Network (FNN)}\label{fig5}
\end{figure}

Traditional time series techniques rely on statistical methods, such as ARIMA, exponential smoothing, and FLM. These methods have been widely used for decades and have proven to be effective in many applications. However, they have some limitations, such as being sensitive to the assumptions made about the data and the need for manual feature engineering. One of the more popular deep learning models for time series prediction is Functional Neural Networks (FNN) \cite{Wang2020ANF, Rossi2002FunctionalDA}, that have been shown to outperform traditional time series models in many applications due to their ability to handle complex and nonlinear relationships in the data. This is particularly useful in applications where the relationships between variables are not well understood or there are many interacting factors. A Functional Neural Network (FNN) \cite{Rao2021NonlinearFM, Rao2021ModernNF}  as shown in Figure \ref{fig5}, is composed of a series of interconnected continuous neurons that are designed to process functional data. The input layer of the network takes in functional data, and each continuous neuron given by Equation \ref{e3} in the continuous hidden layer performs a non linear transformation on the input data or values coming in from the previous layer. The output layer of the network produces a functional output that can be used for prediction or forecasting.

\begin{equation}
\begin{split}
H^{(r)}_{(l)}(s)&=\sigma \Big(b^{(r)}_{(l)}(s) + \sum_{j=1}^{J} \int w^{(r,j)}_{(l)}(s,t)H^{(r,j)}_{(l-1)}(t)dt \Big), \label{e3}
\end{split}
\end{equation}

where $l$ indicates the layer number and $r,j$ are the neuron number, $l=1,2,3,...,L$, $H^{(r)}_{(0)}(s)=X_{r}(s)$ (input time series), $H^{(r)}_{(L)}(s)=\widehat{Y_{r}(s)}$ (output time series), $\sigma(\cdot)$ is a non-linear activation function,
$b^{(r)}_{(l)} \in \mathcal{L}^2(\mcT)$ is the unknown intercept function, $ w^{(r,j)}_{(l)} \in \mathcal L^2(\mcT \times \mcT)$ is the bivariate parameter function for the $r^{th}$ continuous neuron in the $l^{th}$ hidden layer coming from the $j^{th}$ continuous neuron of the $(l-1)^{th}$ hidden layer.

The architecture of FNN can vary widely, depending on the specific application and the complexity of the data. The training typically involves backpropagation through functional derivatives or basis expansion, a method that adjusts the weights of the network to minimize the error between the predicted output and the actual output. The choice of loss function and optimization algorithm also impacts the performance of the network.

\subsection{Model settings}

\subsubsection{Hyperparameter Search}
\begin{table}[]
\begin{tabular}{|l|l|l|}
\hline
Model &
  Selected Features &
  Total No. of Features \\ \hline
LSTM &
  \begin{tabular}[c]{@{}l@{}}Wind Speed\\    \\ Generator Speed\\    \\ Rotator Blade Pitch Angle*Generator Speed\\    \\ Rotator Blade Pitch Angle*Wind Speed Turbulences\end{tabular} &
  4 \\ \hline
FNN &
  \begin{tabular}[c]{@{}l@{}}Wind Speed\\    \\ Generator   Speed\\    \\ Wind   Speed*Generator Speed\\    \\ Wind Speed   Turbulences\end{tabular} &
  4 \\ \hline
\end{tabular}
\caption{Selected Features for LSTM and FNN}
\label{tab1}
\end{table}

To obtain robust power prediction results over all wind turbines, we have conducted an exhaustive hyperparameter search in a grid search manner to understand the performance of the model settings over the dataset with different feature selections. The experiments were conducted separately for two network types: LSTM and FNN. In the end, we have settled on a small set of selected features which are described in Table \ref{tab1} for the two model types.

\begin{table}[]
\footnotesize
\begin{tabular}{|l|l|l|l|l|l|}
\hline
Model &
  Activation Function &
  Loss Error &
  Output Cutoff &
  Learning Rate &
  Network Size \\ \hline
LSTM &
  ReLU &
  MSE &
  Hard cutoff &
  0.001 &
  \begin{tabular}[c]{@{}l@{}}Hidden Size = 2\\    \\ Common Channel = 24\end{tabular} \\ \hline
FNN &
  ELU &
  MSE &
  Sigmoid &
  0.01 &
  \begin{tabular}[c]{@{}l@{}}Hidden Size   = 1\\    \\ Common   Channel = 20\\    \\ Common Size   = 40\end{tabular} \\ \hline
\end{tabular}
\caption{Hyperparameter Settings for LSTM and FNN}
\label{tab2}
\end{table}

For the hyperparameters, we have not only investigated different model sizes but also tested different activation functions (i.e., tanh, ReLU, LeakyReLU, ELU, Sigmoid), learning rates, and error criterions (i.e., MSE, MAE). In addition, wind turbine power output is generally limited to a maximum and minimum possible power output and our models were often over- or underpredicting these limits. Therefore, we have also experimented with different output cutoff methods such as tanh, sigmoid, and a hard cutoff at the minimum and maximum. For the experiments presented in this study, we set our hyperparameters as stated in Table \ref{tab2}.

% \textcolor{red}{adjust table below}

\subsubsection{Ensemble Method}
Ensemble methods involve the amalgamation of multiple machine learning models', a strategy proven to elevate predictive performance, and the ability to combat overfitting, a prevalent issue in machine learning \cite{Liu2019IntelligentWT, Sagi2018EnsembleLA}. Ensemble methods have gained a lot of popularity in recent years for different time series tasks \cite{Lee2023AnEO, Phyo2021HybridED}. Amalgamations can iron out inconsistencies and errors inherent to individual models, reducing bias and variance in the final predictions and leading to enhanced generalization capabilities even in the face of noisy data and outliers. Ensembles can be achieved through a variety of techniques, including employing different algorithms \cite{zhou2012ensemble}, training on distinct data subsets \cite{tian2021rase}, and utilizing varying feature selection strategies \cite{FeatSelectEnsemble}. By virtue of their diversity in model composition, ensembles often produce superior, more reliable prediction across various scenarios.

\subsubsection{Deterioration Detection Approach}
In this study, we leverage the strengths of ensemble methods by amalgamation of the prediction outputs of our separately trained models for FNN and LSTM with an equal weighting of 0.5. Additionally, we investigate a wind turbine equipment health assessment method that leverages the prediction outputs of the ensemble model and compares it to the real outputs by calculating the root mean squared error (RMSE) and the root mean squared percentage error (RMSPE) that are defined in Equation \ref{ermse} and Equation \ref{ermspe} respectively.

% \textcolor{red}{add equation for RMSE and RMSPE}

\begin{equation}
RMSE = \sqrt{\frac{1}{n}\Sigma_{i=1}^{n}{\Big({{Y_{i}} -\widehat{Y_{i}}}\Big)^2}}\label{ermse}
\end{equation}

\begin{equation}
RMSPE = \sqrt{\frac{\Sigma_{i=1}^{n}{\Big({{Y_{i}} -\widehat{Y_{i}}}\Big)^2}}{\Sigma_{i=1}^{n}{{{Y_{i}}}}}}
\label{ermspe}
\end{equation}

\begin{figure}[H]
\includegraphics[width=13.5 cm]{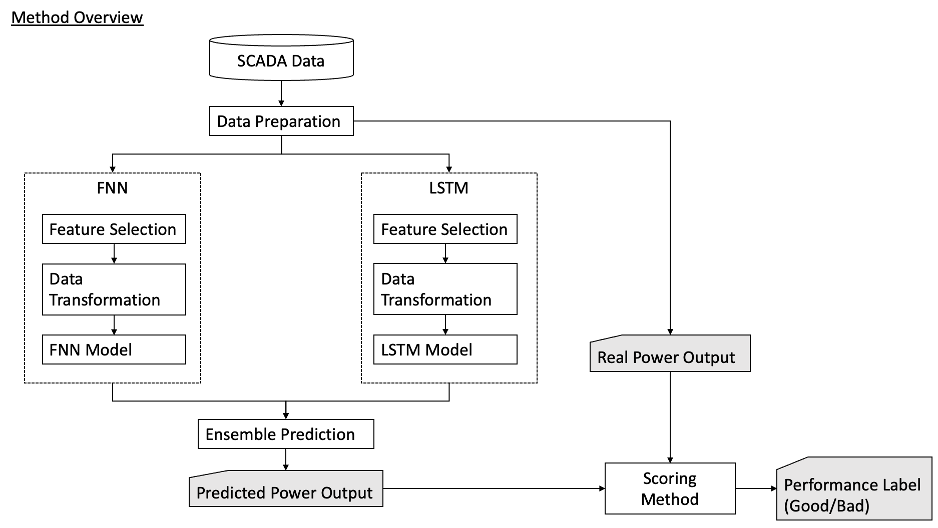}
\caption{Deterioration Detection Method Overview}\label{fig6}
\end{figure}

Our expectation is that the prediction model is trained on the good timeline. If the measured prediction errors increase then it means that the wind turbine power output diverges from the expected power output and the wind turbine performance has degraded.
%from the time series prediction model output, where the model is trained on known good performing time periods of a wind turbine. 
We can calculate RMSE and RMSPE cutoff limits based on the validation dataset for the good timeline. With these cutoff limits, we can detect a degradation in performance of the wind turbine from newly arriving data by comparing predicted and true power output values. An overview of the whole proposed method is shown in Figure \ref{fig6}.

%%%%%%%%%%%%%%%%%%%%%%%%%%%%%%%%%%%%%%%%%%
\section{Results}

\subsection{Prediction Models}

\begin{figure}[H]
\includegraphics[width=13.5 cm]{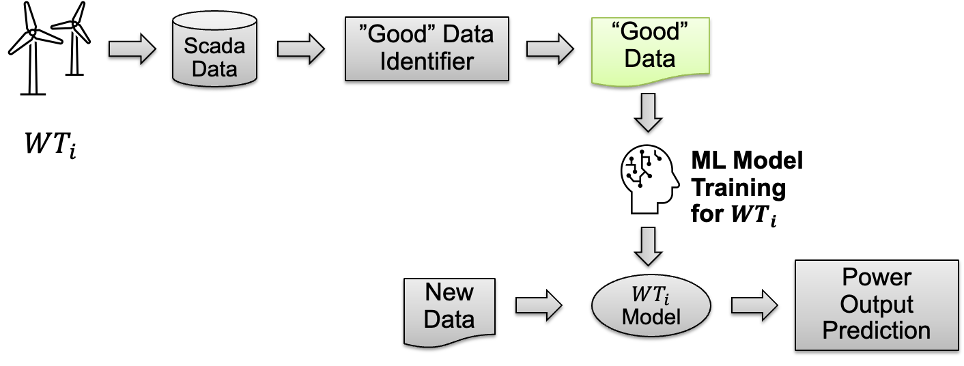}
\caption{Workflow for wind turbine power output prediction}\label{fig7}
\end{figure}  

Figure \ref{fig7} shows the overall flow of our proposed prediction method. In the following, we present the obtained results for prediction and deterioration detection. We have conducted many trial and error studies, where we experimented with LSTM and FNN time series modeling and compared the prediction outputs based on the RMSE. For these studies, we separated the data into “good” and “bad” performing time periods based on domain knowledge obtained from the data provider. A power output prediction model for each wind turbine was trained based on it's “good” data timeline and then tested on both a subset of “good” and “bad” data timelines.

\begin{figure}[H]
\includegraphics[width=8.5 cm]{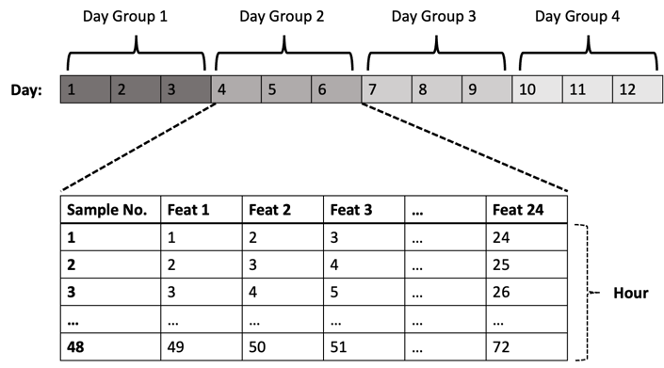}
\caption{Sliding window approach for 3-day time periods}\label{fig8}
\end{figure}  

The training, validation, and test data for the good timeline are created based on randomly ordered three-day time windows to account for different seasons in the training data. To increase the number of samples in the training, validation, and test data, we have used a sliding window approach as depicted in Figure \ref{fig8}. First, we assign hourly resampled data to a 3-day long day group and then apply a sliding window approach within each 3-day group. This makes sure there is no data leakage. With this approach, we can increase the number of total samples while enabling the random assignment the training, validation, or test dataset. For our first prediction evaluation, we created the test dataset for the bad timeline in a similar manner, and we have aligned the randomization over all wind turbines to compare the same timelines in the same training, validation or test categories.

\begin{figure}[H]
\includegraphics[width=13.5 cm]{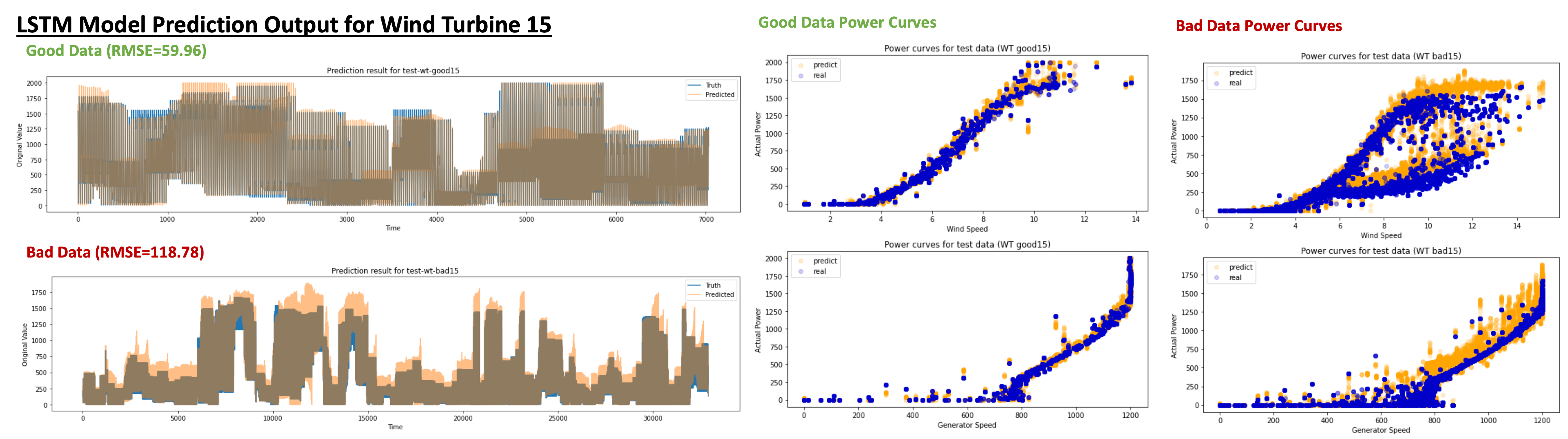}
\caption{LSTM Model prediction output for both good and bad data periods}\label{fig9}
\end{figure}  

\begin{figure}[H]
    \includegraphics[width=1\linewidth]{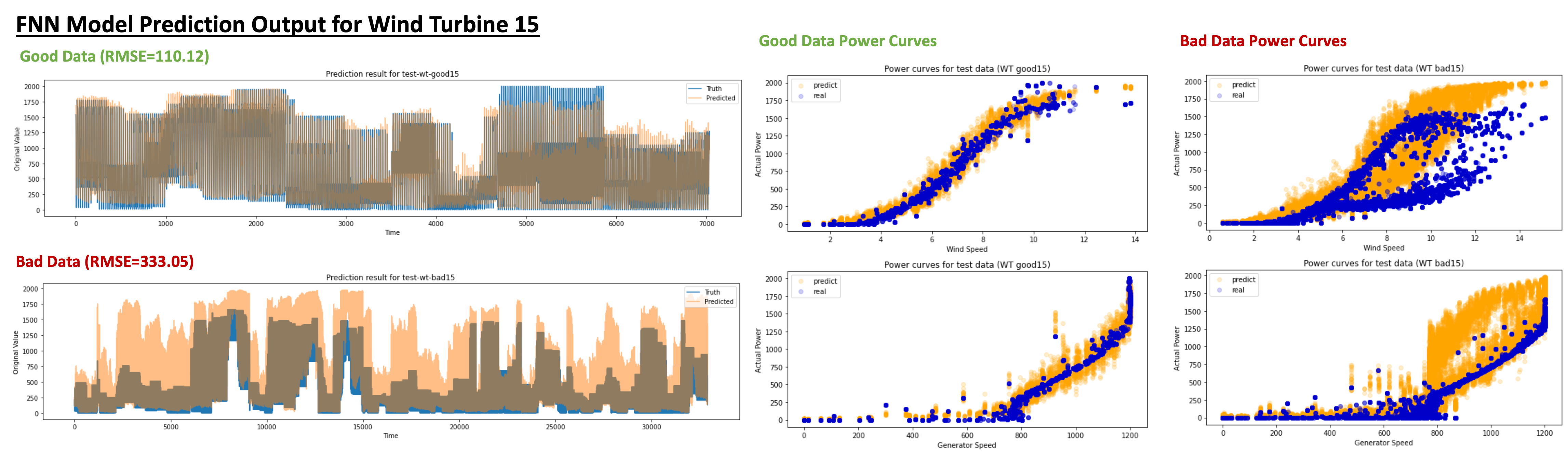}
    \caption{FNN Model prediction output for both good and bad data periods}
    \label{fig9b}
\end{figure}

An example of how such power output predictions would look like for a sample wind turbine no. 15 using LSTM and FNN model is shown in Figure \href{}{\ref{fig9}} and Figure \ref{fig9b} respectively  for both “good” and “bad” test datasets. We can clearly observe that the prediction model is often overpredicting the power output for the bad timeline datasets, which indicates that the maximum power output could be potentially higher and there is an issue with the wind turbine performance. The poor performance can also be confirmed on the bad timeline data's power curves for the wind and generator speed (right side of Figure \ref{fig9} and Figure \ref{fig9b}), as it deviates away from it's sigmoid behaviour. Comparing the predictions of LSTM and FNN, we observe that LSTM achieves a higher prediction accuracy on both good and bad test datasets whereas FNN offers more separation between them.

\begin{figure}
    \centering
    \includegraphics[width=0.55\linewidth]{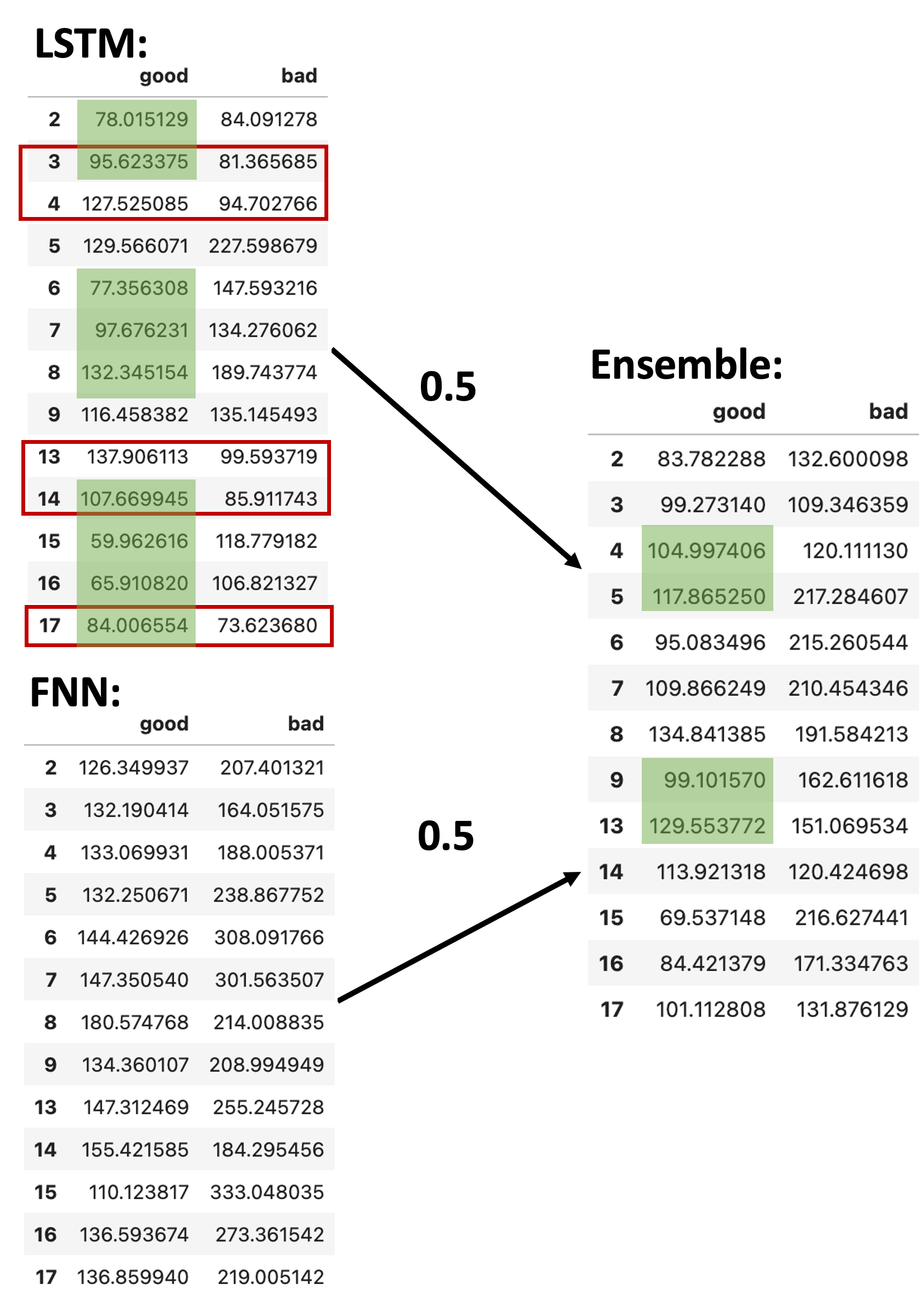}
    \caption{Ensemble Prediction Performance Comparison}
    \label{fig10}
\end{figure}

This observation can be further confirmed in Figure \ref{fig10} that shows the RMSE for predictions of three different models LSTM, FNN and an ensembled prediction model where LSTM and FNN predictions are accounted for with equal weights. Overall, we can conclude that the LSTM model performs better on all 13 wind turbine indices for both good and bad test datasets in comparison to FNN. However, the LSTM model also performs suspiciously well on the bad test datasets compared to the good datasets for the wind turbines no. 3, 4, 13, 14 and 17 (highlighted in red in Figure \ref{fig10}). The model might be overfitting. Since we want to leverage the prediction accuracy as an indicator of performance deterioration, we need a high performance accuracy on good timelines and a poor performance accuracy on bad timelines, the latter part is provided by FNN. 
We achieve this by combining the LSTM and FNN model predictions in an equally weighted ensemble shown on the right side of  Figure \ref{fig10}. The green highlights indicate the best performing model for each of the wind turbine on the good timeline test datasets. The ensemble model achieves a similar performance on good timeline datasets and even outperforms LSTM prediction accuracy in four cases (for wind turbine no. 4, 5, 9 and 13). We also achieve a bigger gap between prediction accuracy's for the good timeline and bad timeline datasets. For the ensemble method, no wind turbine prediction model performs better on the bad timeline dataset compared to the good timeline dataset.

\begin{figure}
    \centering
    \includegraphics[width=1\linewidth]{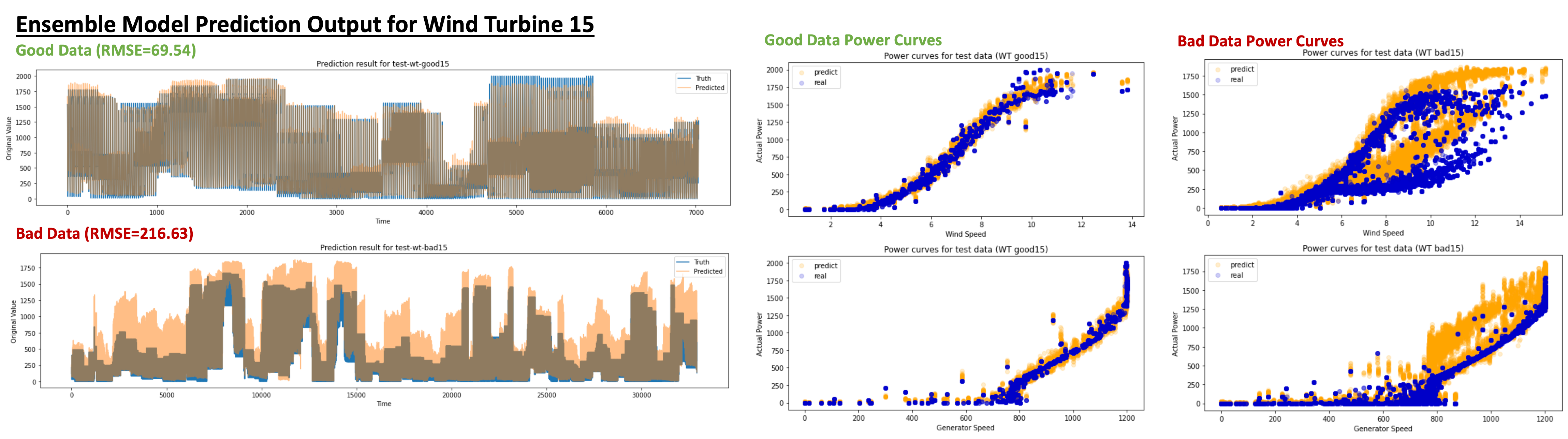}
    \caption{Ensemble Model prediction output for both good and bad data periods}
    \label{fig11}
\end{figure}

\begin{figure}
      \centering
      \includegraphics[width=0.75\linewidth]{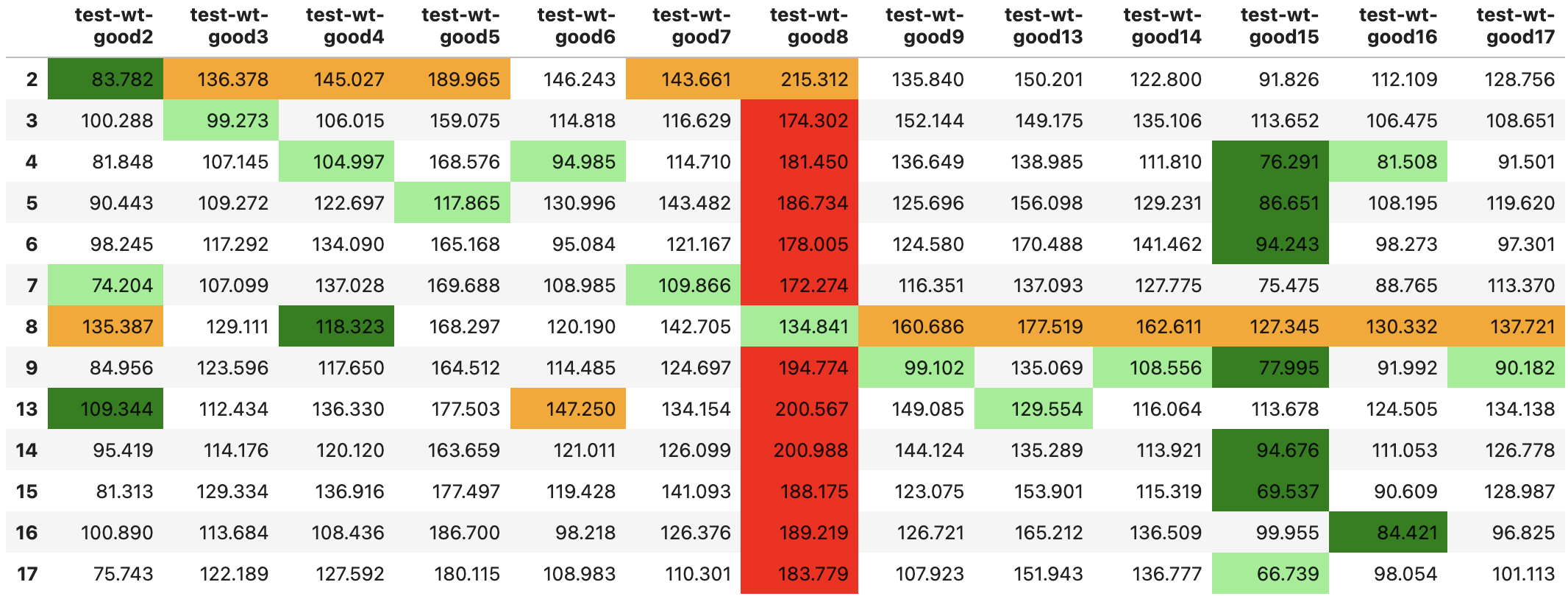}
      \caption{Ensemble Prediction Results over all good timeline test datasets}
      \label{fig12}
  \end{figure}  

Figure \ref{fig11} shows the prediction performance for our previous example wind turbine no. 15 for the good timeline and bad timeline test datasets. The performance stays robust for the good timeline dataset but prediction performance deterioration for the bad timeline dataset is now much clearer. In Figure \ref{fig12}, we show the prediction results (RMSE) for the “good timeline” test dataset over all wind turbines for the ensemble prediction model. The first column of the table is the wind turbine no. on whose training data the wind turbine prediction model was built. Each wind turbine prediction model was used to not only predict the power output for its own test dataset but for all other wind turbines’ “good timeline” test datasets as well. The obtained results are interesting, there's an overall tendency we can observe that each wind turbine is performing comparably well on its own test dataset as shown by the light green highlights indicating the minimum performance error of each column that often aligns with a wind turbine's own prediction model. Also, we can observe that performance results are mostly worst for wind turbine 8 (red highlight indicates max performance error of each row) and best for wind turbine 15 (dark green highlight indicates minimum performance error of each row). Wind turbine 8 might be affected by factors that are not considered in the selected features (or some unknown external features) making the power prediction more difficult while wind turbine 15 is the most straight forward to predict.

\begin{figure}
    \centering
    \includegraphics[width=0.75\linewidth]{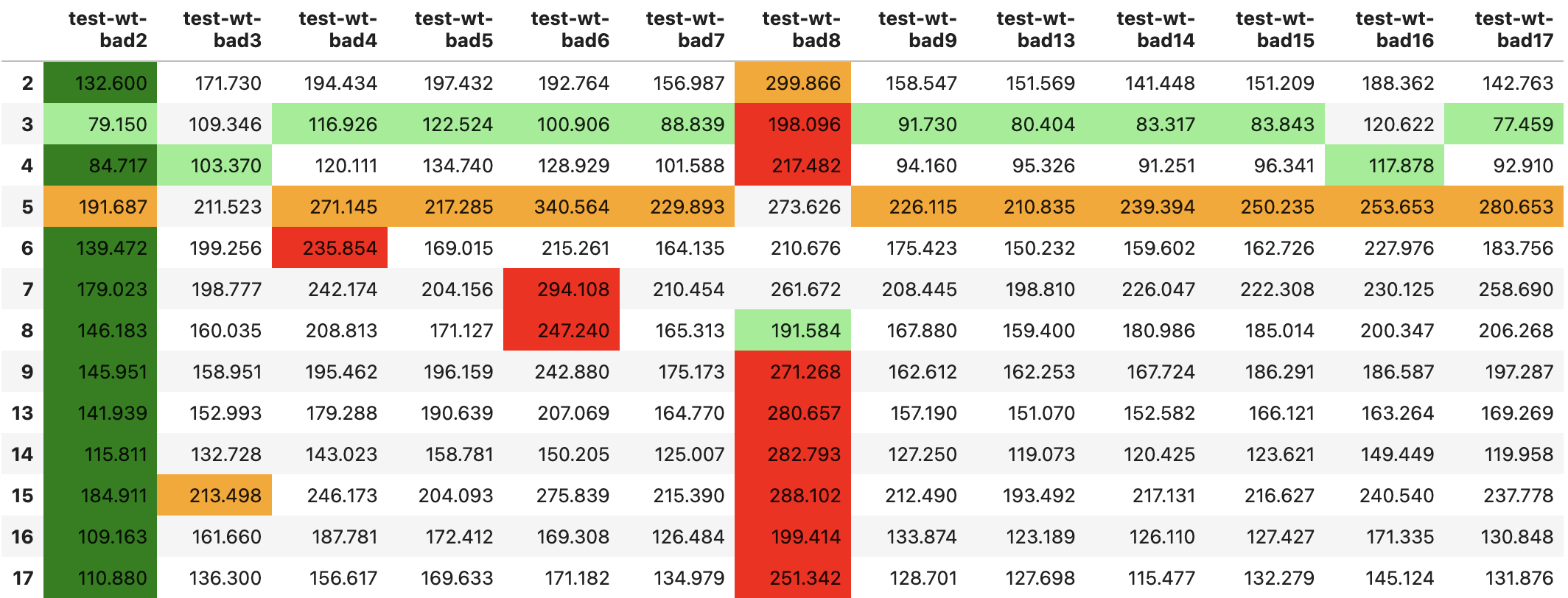}
    \caption{Ensemble Prediction Results over all bad timeline test datasets}
    \label{fig13}
\end{figure}

In Figure \ref{fig13}, we can see the ensemble model prediction results (RMSE) for the bad timeline test dataset over all wind turbines. The first column of the table is the wind turbine no. on whose good timeline training dataset the prediction model was built. We used the prediction models for each wind turbine to not only predict the power output for its own test dataset but for all other wind turbines’ “bad timeline” test datasets. Overall, we can observe that each wind turbine is performing  worse compared to the “good timeline” test datasets in Figure \ref{fig12}. Also, we can observe that prediction performance results are worst for wind turbine 8 and best for wind turbine 2 regardless of the chosen prediction model. On the other hand, the prediction model for wind turbine no. 3 achieves the overall best prediction performance (light green highlights for the best entry in each column) and wind turbine no. 5 the worst (orange highlights for worst entry in each column).

In summary, Figure \ref{fig12} showed us that the best prediction model for a wind turbine is based on its own training data. Results of Figure \ref{fig13} on the other hand showed that there might be one wind turbine prediction model that is overachieving on the bad timeline of another wind turbine. Therefore, we can't use any one of the wind turbine’s prediction model for other wind turbines' performance prediction. This supports our claim that each wind turbine (or equipment) is unique and has different feature settings, hence they can't be considered as homogeneous. Therefore, we have decided to predict power output for each wind turbine separately. 

%The next step is then to decide whether FNN or LSTM should be chosen as the preferred model. We have compared the performance results on the “good timeline” and “bad timeline” test data for LSTM, FNN, and an ensemble of both as seen Figure . Interestingly, for the “bad timeline”, LSTM is often performing surprisingly well opposite to our expectation of bad prediction performance. This needs further investigation. We can observe that LSTM shows slightly better performance over FNN for the “good timeline” but this can be often further improved by creating an ensemble prediction of the two. We can also see a clear difference in performance between the good timeline and bad timeline for the ensemble method. The ensemble helps us to take the strength of both methods and make a more robust and precise prediction of the power output.

\subsection{Deterioration detection}
For deterioration detection, we have to find the balance between good performance on the good timeline and bad performance on the bad timeline. This needs to be achieved over a range of wind turbines (or other equipment) individually. 
%without fine-tuning prediction models for each equipment individually. 
The ensemble prediction method shows the most robust performance and therefore is our time series prediction method of choice.

\begin{figure}[H]
\includegraphics[width=13.5 cm]{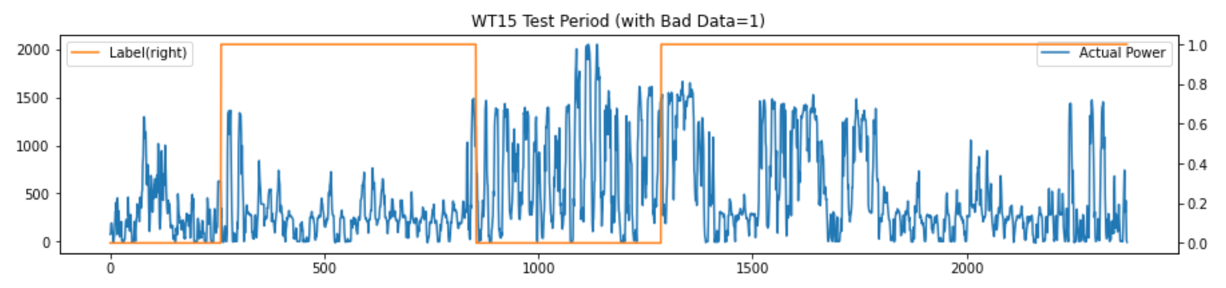}
\caption{Test Data for Wind Turbine 15 with Labels for deterioration detection}\label{fig15}
\end{figure}  

For deterioration detection, we use a continuous time period with good and bad timelines to test the performance of our approach. Figure \ref{fig15} is showing an example test timeline for wind turbine no. 15 that has good and bad timelines. The bad timelines are labeled by the value 1 indicated by the orange line on the secondary y-axis. All labels were manually assigned according to the available information from the data provider. 

\begin{figure}[H]
\includegraphics[width=13.5 cm]{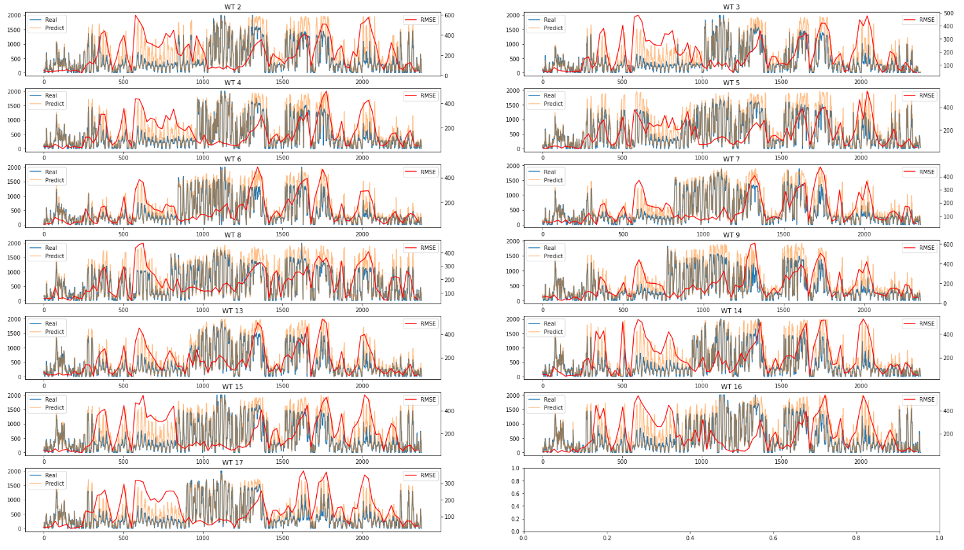}
\caption{Power Output Prediction Results for all Wind Turbines}\label{fig16}
\end{figure}

Figure \ref{fig16} shows the power output prediction results that are obtained for each wind turbine based on the workflow presented in Figure \ref{fig7} using the ensemble prediction method. We can observe that the RMSE (red line, scale on secondary y-axis), calculated for 24-hour windows, is generally lower in the very beginning and in the middle of the presented test timeline. This observation matches with the time periods that we have identified as “good timelines” based on our domain knowledge.

\begin{figure}[H]
\centering
\includegraphics[width=13.5 cm]{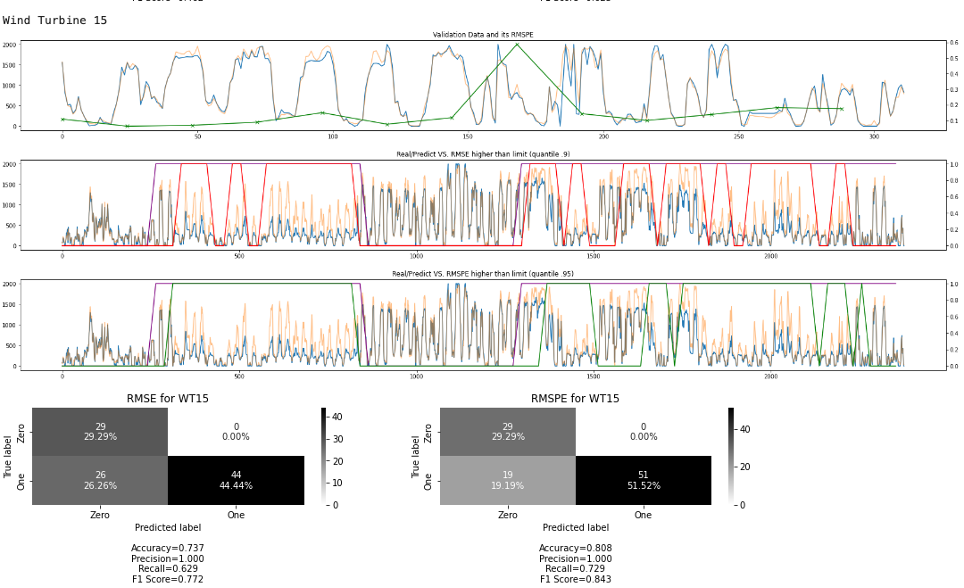}
\caption{Deterioration Detection for Wind Turbine 15}\label{fig17}
\end{figure}

% these para needs editing. We can discuss together and make the edits

% \textcolor{red}{we will to edit the para below: already edited it a bit, let's recheck together}

To detect deterioration in the performance of a wind turbine, we want to establish a prediction model performance baseline using RMSE or RMSPE metrics leveraging validation data, i.e., a known good timeline dataset that wasn't used during model training. This baseline can then be compared to the prediction performance of a test dataset where the real wind turbine output performance is unknown.
%Then, we have conducted deterioration detection by comparing the prediction performance (RMSE, RMSPE) of the test data to the one of the validation data. 
Figure \ref{fig17} shows an example of this process for wind turbine no. 15. The first graph shows the validation data and its RMSPE (green line on secondary y-axis). Once RMSPE value is calculated for each 24-hour window hence the green line is cut off early in the plot and RMSE can be calculated in the same manner. Then, we use the obtained RMSE/RMSPE values to calculate value cutoff limits using simple statistics. %for future RMSE/RMSPE values as considered in the test data. 
For this research, the cutoff limits are calculated individually for each wind turbine prediction model using the 90\% percentile RMSE value and the 95\% percentile RMSPE value based on the validation data. Thus, dynamic cutoff limits can be calculated based on the prediction quality for each wind turbine. The cutoff limits are then used to identify if an RMSE/RMSPE value observed for the test data suggests a wind turbine performance deterioration or not. Examples of this can be seen in Figure \ref{fig17} for wind turbine no. 15, where the second plot shows the RMSE based deterioration labels in red and true labels in purple. The third plot shows similar information for RMSPE with true labels in purple and identified labels in green. The two plots in the last row of Figure \ref{fig17} show the confusion matrices for RMSE and RMSPE respectively. The plots give us insights into the performance deterioration classification accuracy of the two metrics. Our proposed performance deterioration approach performs better when we use the RMPSE metric. We observe 100\% precision, meaning that if the wind turbine is in a good state, we are always able to classify this correctly. There are cases when the method is unable to detect deterioration correctly, especially when the expected power output is in a lower value range or at the beginning/end of a bad performing time period. Performance deterioration might therefore be harder to detect when wind speed and actual power output are low.  We conclude that the impact of deterioration is felt more during times where the expected power output is supposed to be high.

\begin{figure}[H]
\includegraphics[width=13.5 cm]{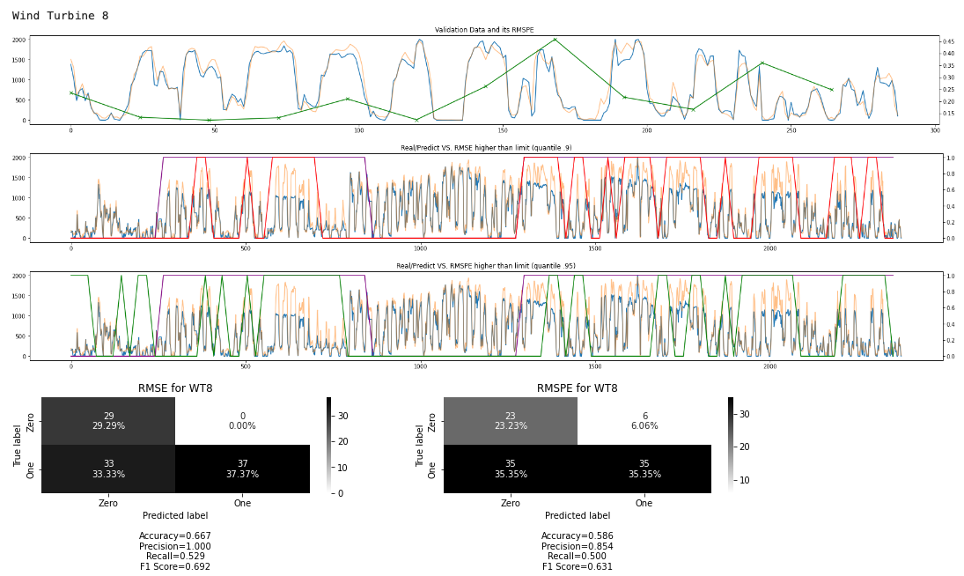}
\caption{Deterioration Detection for Wind Turbine 8}\label{fig18}
\end{figure}  

Figure \ref{fig18} shows a second deterioration detection example for wind turbine no. 8, which we observed previously as the most difficult wind turbine to achieve good prediction results for as shown in Figure \ref{fig12} and Figure \ref{fig13}. For this wind turbine, the classification accuracy for the deterioration task is slightly lower. The observed F1-score is 0.631 for the RMSPE metric and the lower score is observed because less of the bad performing time periods are actually identified as bad and some good performing time periods also have been wrongly identified as bad. 
%\textcolor{red}{All the above calculated classification metrics are based on the binary use case where we are mostly interested in the true labels, i.e., bad performance labels, and don’t consider the correct identification of “good timeline”.} Therefore, 
In the following  Table \ref{tab3}, we calculate the overall weighted F1 classification score to account for the imbalance in the classes for each wind turbine to understand the overall performance of our proposed method. % as shown in Table \ref{tab3}.
In addition to the RMSE and RMSPE metrics, we also added a mixed metric version where a bad time period is identified if either RMSE or RMSPE value cutoff limits are triggered. Overall, we observe the highest classification performance for the mixed metric that outperforms the other two approaches in 10 out of 13 cases. In the other cases, the performance is equal or slightly worse compared to the RMSPE approach. The RMSE approach never performs the best and only beats the RMSPE approach for wind turbine no. 3 and 8. The reason why the relative percentage error like RMSPE works better is that the bad performing time periods include expected low value power outputs that are harder to detect when only using statistically calculated cutoff limits based on an absolute value. Therefore, our conclusion is to use the mixed approach for getting the best performance in deterioration detection.

%\textcolor{red}{add few lines discussing the table}

\begin{table}[H]
\centering
\begin{tabular}{|l|l|l|l|} \hline
Wind Turbine & RMSE & RMSPE & Mixed \\ \hline
2 & 0.661 & 0.737 & \textbf{0.783} \\
3 & 0.710 & 0.704 & \textbf{0.775} \\
4 & 0.537 & 0.747 & \textbf{0.793} \\
5 & 0.407 & 0.516 & \textbf{0.634} \\
6 & 0.549 & \textbf{0.890} & 0.878 \\
7 & 0.490 & \textbf{0.876} & \textbf{0.876} \\
8 & 0.675 & 0.600 & \textbf{0.720} \\
9 & 0.526 & \textbf{0.823} & 0.812 \\
13 & 0.457 & 0.720 & \textbf{0.768} \\
14 & 0.485 & 0.605 & \textbf{0.648} \\
15 & 0.748 & 0.816 & \textbf{0.873} \\
16 & 0.717 & 0.754 & \textbf{0.757} \\
17 & 0.573 & 0.883 & \textbf{0.892} \\ \hline
\end{tabular}
\centering
\caption{Weighted F1-Score Results for RMSE, RMSPE, and Mixed Metric}
\label{tab3}
\end{table}

%%%%%%%%%%%%%%%%%%%%%%%%%%%%%%%%%%%%%%%%%%

\section{Discussion}
In the following section, we will discuss the results and benefits of the proposed method. In summary, we have proposed a wind turbine output prediction modeling and deterioration detection method that takes data from a wind turbine and then trains a prediction model to conduct deterioration detection on newly observed data. While the approach processes each wind turbine separately, the workflow is automatized in a manner where fine-tuning to the needs of each turbine is not necessary.

The results for the power output prediction show that not all wind turbines achieve the same prediction performance due to differences in their feature settings even though the turbines were physically placed in a similar environment within a short distance from each other. It is therefore important to create separate prediction models for each wind turbine. Furthermore, our previous discussion based on Figure \ref{fig10}  showed that it is hard to choose a specific time series prediction method, LSTM or FNN, over another for all the wind turbines and an ensemble is a good approach to stabilize the prediction results achieving good prediction accuracy on the good timelines and distinguishably worse prediction performance on the bad timelines.

For the deterioration detection, we have shown an overview of all wind turbines' power output prediction performance plotted for a continuous test dataset in Figure \ref{fig16}. The overall periods of good and bad performance are similar for the deterioration detection of all wind turbines showing that the wind turbines are operating under similar conditions. On the other hand, our experiments showed the uniqueness of each wind turbine. This makes us confident that there is value in using our ensemble model for prediction to detect deterioration of wind turbine individually. %especially since we have domain knowledge that there were some deterioration issues due to dust collection during the considered test time frame. 
% We have manually assigned labels to the whole test period and compared our automatically detected labels to the manual labels calculating classification scores.
The overall F1-scores as shown in Table \ref{tab3} are between 0.634 and 0.892 for all wind turbines for a mixed error metrics. %\textcolor{red}{These results can be considered good, given that the assumed “true labels” are manually assigned by us and do not necessarily represent the true nature completely.} 
In addition, it is not necessary for deterioration detection to identify all labels correctly. If we can keep the number of false positives low during “good timeline” data periods while detecting continuous phases of bad performance during “bad timeline” periods, it will be possible for a wind turbine operator to use these results to improve overall wind turbine power output in a quick manner. Even a fraction of an improvement will result in an improved business value.

Our work can identify deteriorating performance in wind turbines therefore enabling better planning for the maintenance and repairs of wind turbines. The proposed deterioration detection approach can also be applied to other equipment and help improve the performance of those industry machinery. Running machines at lower efficiency leads to faster wear and tear as well as revenue loss, and our work supports to mitigate these risks. The obtained results are promising and it will be interesting to apply our work to other machines/equipment. The accurate prediction of power output over a group of wind turbines using the same method is challenging because of the differences in external and internal factors and how they affect the wind turbines. Currently, data is limited, and training/validation has to be conducted on a relatively small dataset, especially for the second part of deterioration detection. More experiments need to be done over a longer horizon to understand the performance better.

% \textcolor{red}{we will edit below as needed}

% we will need to validate with atria on this
We shared our comprehensive analysis with Atria Power, the owner of the wind turbine farm, and they found it to be valuable. The insights derived from our analysis align closely with the operational dynamics of their wind turbines, providing them with actionable intelligence to enhance their operations. Atria Power is excited about the prospect of implementing these insights and is eager to explore how our findings can be integrated into their existing processes. They perceive a significant potential in this collaboration, recognizing the opportunity to improve efficiency, reliability, and overall performance of their wind turbine operations.

% %%%%%%%%%%%%%%%%%%%%%%%%%%%%%%%%%%%%%%%%%%
\section{Conclusions}

We have demonstrated how to efficiently predict the power output of a wind turbine using SCADA data and then to leverage these predictions to detect possible performance deterioration. This analysis can be extended to any machine and helps to proactively maintain such equipment in a healthy state, to run them optimally and to get the best performance while maximizing profit for the business/company.

Our presented results showed that we were able to predict deterioration for all wind turbines where the classification F1-score was between 0.634 and 0.892. We achieved this with an automatized workflow that can create a prediction model based on training data, predict outputs for a validation dataset to calculate performance cutoff limits, and then label newly observed test data as good or bad performing. Although we did some initial trial and error to achieve a stable prediction performance over all wind turbines, we refrained from fine-tuning/specializing prediction models to the needs of specific wind turbines or do anything that might make application in a real-world business scenario difficult.
% We also explored data preprocessing methods that help to increase the amount of available training data for business scenarios where data is limited.
% Indeed, the overall data time period that was available for our study is approximately half a year, so we were not able to evaluate prediction performance over a bigger time horizon of one year or more yet and this remains future work.

For future work, we aim to broaden the scope of our research by exploring additional deep neural networks and time series prediction approaches to further enhance our ensemble model. Extending our analysis to encompass broader class of machinery will allow us to generalize our findings and contribute to a more comprehensive understanding of predictive maintenance. Additionally, incorporating a Functional Basis Neural Network with Fourier or wavelet functions holds promise for improving modeling results, warranting further investigation. Also, as next steps, we plan to incorporate our work to more Atria Power Wind Farms. Finally, expanding our time scope beyond six months will enable us to capture and account for all seasonalities, other kinds of bad states, providing a more comprehensive assessment of performance over time.

\bibliographystyle{apalike}
\bibliography{ref}

\begin{thebibliography}{}

\bibitem[Backhus and Kono, 2022]{backhus2022cooling}
Backhus, J. and Kono, Y. (2022).
\newblock Cooling power consumption dependency simulation for real-world data center data.
\newblock In {\em 2022 International Conference on Software, Telecommunications and Computer Networks (SoftCOM)}, pages 1--6. IEEE.

\bibitem[bo~Jin et~al., 2019]{Jin2019PredictionFT}
bo~Jin, X., Yu, X.-H., Wang, X., ting Bai, Y., Su, T., and Kong, J. (2019).
\newblock Prediction for time series with cnn and lstm.

\bibitem[Dolara et~al., 2015]{dolara2015comparison}
Dolara, A., Leva, S., and Manzolini, G. (2015).
\newblock Comparison of different physical models for pv power output prediction.
\newblock {\em Solar energy}, 119:83--99.

\bibitem[Ferraty and Romain, 2011]{Ferraty2011TheOH}
Ferraty, F. and Romain, Y. (2011).
\newblock The oxford handbook of functional data analysis.
\newblock {\em Oxford Handbooks Online}.

\bibitem[Gernaat et~al., 2021]{Gernaat2021ClimateCI}
Gernaat, D.~E., de~Boer, H.~S., Daioglou, V., Yalew, S.~G., M{\"u}ller, C., and van Vuuren, D.~P. (2021).
\newblock Climate change impacts on renewable energy supply.
\newblock {\em Nature Climate Change}, 11:119 -- 125.

\bibitem[Guan et~al., 2014]{FeatSelectEnsemble}
Guan, D., Yuan, W., Lee, Y.-K., Najeebullah, K., and Rasel, M.~K. (2014).
\newblock A review of ensemble learning based feature selection.
\newblock {\em IETE Technical Review}, 31(3):190--198.

\bibitem[Han et~al., 2019]{Han2019ARO}
Han, Z., Zhao, J., Leung, H., Ma, K., and Wang, W. (2019).
\newblock A review of deep learning models for time series prediction.
\newblock {\em IEEE Sensors Journal}, 21:7833--7848.

\bibitem[Hochreiter and Schmidhuber, 1997]{Hochreiter1997LongSM}
Hochreiter, S. and Schmidhuber, J. (1997).
\newblock Long short-term memory.
\newblock {\em Neural Computation}, 9:1735--1780.

\bibitem[Jin et~al., 2021]{Jin2021ConditionMO}
Jin, X., Xu, Z., and Qiao, W. (2021).
\newblock Condition monitoring of wind turbine generators using scada data analysis.
\newblock {\em IEEE Transactions on Sustainable Energy}, 12:202--210.

\bibitem[Koch et~al., 2015]{Koch2015TheIO}
Koch, H., V{\"o}gele, S., Hattermann, F.~F., and Huang, S. (2015).
\newblock The impact of climate change and variability on the generation of electrical power.
\newblock {\em Meteorologische Zeitschrift}, 24:173--188.

\bibitem[Kokoszka and Reimherr, 2017]{Kokoszka2017IntroductionTF}
Kokoszka, P. and Reimherr, M.~L. (2017).
\newblock Introduction to functional data analysis.

\bibitem[Lee et~al., 2023]{Lee2023AnEO}
Lee, X.~Y., Kumar, A., Vidyaratne, L.~S., Rao, A.~R., Farahat, A.~K., and Gupta, C.~R. (2023).
\newblock An ensemble of convolution-based methods for fault detection using vibration signals.
\newblock {\em 2023 IEEE International Conference on Prognostics and Health Management (ICPHM)}, pages 172--179.

\bibitem[Lindemann et~al., 2021]{Lindemann2021ASO}
Lindemann, B., M{\"u}ller, T., Vietz, H., Jazdi, N., and Weyrich, M. (2021).
\newblock A survey on long short-term memory networks for time series prediction.
\newblock {\em Procedia CIRP}, 99:650--655.

\bibitem[ling Wang et~al., 2016]{Wang2016FunctionalDA}
ling Wang, J., Chiou, J.-M., and M{\"u}ller, H.-G. (2016).
\newblock Functional data analysis.

\bibitem[Liu et~al., 2019]{Liu2019IntelligentWT}
Liu, Y., Cheng, H., Kong, X., bin Wang, Q., and Cui, H. (2019).
\newblock Intelligent wind turbine blade icing detection using supervisory control and data acquisition data and ensemble deep learning.
\newblock {\em Energy Science \& Engineering}, 7:2633 -- 2645.

\bibitem[Maldonado-Correa et~al., 2020]{MaldonadoCorrea2020UsingSD}
Maldonado-Correa, J.~L., Mart{\'i}n-Mart{\'i}nez, S., Artigao, E., and G{\'o}mez-L{\'a}zaro, E. (2020).
\newblock Using scada data for wind turbine condition monitoring: A systematic literature review.
\newblock {\em Energies}.

\bibitem[Massidda and Marrocu, 2017]{massidda2017use}
Massidda, L. and Marrocu, M. (2017).
\newblock Use of multilinear adaptive regression splines and numerical weather prediction to forecast the power output of a pv plant in borkum, germany.
\newblock {\em Solar Energy}, 146:141--149.

\bibitem[Narasinh et~al., 2024]{Narasinh2024InvestigatingPL}
Narasinh, V., Mital, P., Chakravortty, N., Mittal, S., Kulkarni, N., Venkatraman, C., Rajakumar, A.~G., and Banerjee, K. (2024).
\newblock Investigating power loss in a wind turbine using real-time vibration signature.
\newblock {\em Engineering Failure Analysis}.

\bibitem[Pandit et~al., 2022]{Pandit2022SCADADF}
Pandit, R.~K., Astolfi, D., Hong, J., Infield, D., and Santos, M. (2022).
\newblock Scada data for wind turbine data-driven condition/performance monitoring: A review on state-of-art, challenges and future trends.
\newblock {\em Wind Engineering}, 47:422 -- 441.

\bibitem[Parmezan et~al., 2019]{Parmezan2019EvaluationOS}
Parmezan, A. R.~S., Souza, V. M.~A., and Batista, G. E. A. P.~A. (2019).
\newblock Evaluation of statistical and machine learning models for time series prediction: Identifying the state-of-the-art and the best conditions for the use of each model.
\newblock {\em Inf. Sci.}, 484:302--337.

\bibitem[Phyo and Byun, 2021]{Phyo2021HybridED}
Phyo, P.~P. and Byun, Y. (2021).
\newblock Hybrid ensemble deep learning-based approach for time series energy prediction.
\newblock {\em Symmetry}, 13:1942.

\bibitem[Port{\'e}-Agel et~al., 2019]{PortAgel2019WindTurbineAW}
Port{\'e}-Agel, F., Bastankhah, M., and Shamsoddin, S. (2019).
\newblock Wind-turbine and wind-farm flows: A review.
\newblock {\em Boundary-Layer Meteorology}, 174:1 -- 59.

\bibitem[Pryor et~al., 2020]{Pryor2020ClimateCI}
Pryor, S.~C., Barthelmie, R.~J., Bukovsky, M.~S., Leung, L.~R., and Sakaguchi, K. (2020).
\newblock Climate change impacts on wind power generation.
\newblock {\em Nature Reviews Earth \& Environment}, 1:627 -- 643.

\bibitem[Ramsay and Silvermann, 1997]{Ramsay1997FunctionalDA}
Ramsay, J.~O. and Silvermann, B.~W. (1997).
\newblock Functional data analysis (springer series in statistics).

\bibitem[Rao and Reimherr, 2021a]{Rao2021ModernNF}
Rao, A.~R. and Reimherr, M.~L. (2021a).
\newblock Modern non-linear function-on-function regression.
\newblock {\em Statistics and Computing}, 33:1--12.

\bibitem[Rao and Reimherr, 2021b]{Rao2021NonlinearFM}
Rao, A.~R. and Reimherr, M.~L. (2021b).
\newblock Nonlinear functional modeling using neural networks.
\newblock {\em Journal of Computational and Graphical Statistics}, 32:1248 -- 1257.

\bibitem[Reiss et~al., 2017]{Reiss2017MethodsFS}
Reiss, P.~T., Goldsmith, J., Shang, H.~L., and Ogden, R.~T. (2017).
\newblock Methods for scalar‐on‐function regression.
\newblock {\em International Statistical Review}, 85:228 -- 249.

\bibitem[Rossi et~al., 2002]{Rossi2002FunctionalDA}
Rossi, F., Conan-Guez, B., and Fleuret, F. (2002).
\newblock Functional data analysis with multi layer perceptrons.
\newblock {\em Proceedings of the 2002 International Joint Conference on Neural Networks. IJCNN'02 (Cat. No.02CH37290)}, 3:2843--2848 vol.3.

\bibitem[Sagi and Rokach, 2018]{Sagi2018EnsembleLA}
Sagi, O. and Rokach, L. (2018).
\newblock Ensemble learning: A survey.
\newblock {\em Wiley Interdisciplinary Reviews: Data Mining and Knowledge Discovery}, 8.

\bibitem[Song et~al., 2020]{Song2020TimeseriesWP}
Song, X., Liu, Y., Xue, L., Wang, J., Zhang, J., Wang, J., Jiang, L., and Cheng, Z. (2020).
\newblock Time-series well performance prediction based on long short-term memory (lstm) neural network model.
\newblock {\em Journal of Petroleum Science and Engineering}, 186:106682.

\bibitem[Tautz-Weinert and Watson, 2017]{TautzWeinert2017UsingSD}
Tautz-Weinert, J. and Watson, S.~J. (2017).
\newblock Using scada data for wind turbine condition monitoring – a review.
\newblock {\em Iet Renewable Power Generation}, 11:382--394.

\bibitem[Theocharides et~al., 2018]{theocharides2018machine}
Theocharides, S., Makrides, G., Georghiou, G.~E., and Kyprianou, A. (2018).
\newblock Machine learning algorithms for photovoltaic system power output prediction.
\newblock In {\em 2018 IEEE International Energy Conference (ENERGYCON)}, pages 1--6. IEEE.

\bibitem[Tian and Feng, 2021]{tian2021rase}
Tian, Y. and Feng, Y. (2021).
\newblock Rase: Random subspace ensemble classification.
\newblock {\em The Journal of Machine Learning Research}, 22(1):2019--2111.

\bibitem[Veers et~al., 2019]{Veers2019GrandCI}
Veers, P.~S., Dykes, K., Lantz, E.~J., Barth, S., Bottasso, C.~L., Carlson, O., Clifton, A., Green, J.~B., Green, P., Holttinen, H., Laird, D.~L., Lehtom{\"a}ki, V., Lundquist, J.~K., Lundquist, J.~K., Manwell, J., Marquis, M., Meneveau, C., Moriarty, P.~J., Munduate, X., Muskulus, M., Naughton, J.~W., Pao, L.~Y., Paquette, J.~A., Peinke, J., Robertson, A.~N., Rodrigo, J.~S., Sempreviva, A.~M., Smith, J.~C., Tuohy, A., and Wiser, R. (2019).
\newblock Grand challenges in the science of wind energy.
\newblock {\em Science}, 366.

\bibitem[Wang et~al., 2022]{Wang2022FaultDA}
Wang, H., Zhao, X., and Wang, W. (2022).
\newblock Fault diagnosis and prediction of wind turbine gearbox based on a new hybrid model.
\newblock {\em Environmental Science and Pollution Research}, 30:24506--24520.

\bibitem[Wang et~al., 2021]{wang2021comparison}
Wang, M., Peng, J., Luo, Y., Shen, Z., and Yang, H. (2021).
\newblock Comparison of different simplistic prediction models for forecasting pv power output: assessment with experimental measurements.
\newblock {\em Energy}, 224:120162.

\bibitem[Wang et~al., 2020]{Wang2020ANF}
Wang, Q., Wang, H., Gupta, C., Rao, A.~R., and Khorasgani, H. (2020).
\newblock A non-linear function-on-function model for regression with time series data.
\newblock {\em 2020 IEEE International Conference on Big Data (Big Data)}, pages 232--239.

\bibitem[Yang et~al., 2014]{Yang2014WindTC}
Yang, W., Tavner, P.~J., Crabtree, C.~J., Feng, Y., and Qiu, Y. (2014).
\newblock Wind turbine condition monitoring: technical and commercial challenges.
\newblock {\em Wind Energy}, 17:673--693.

\bibitem[Zhang et~al., 2019]{Zhang2019DataDrivenMF}
Zhang, W., Yang, D., and Wang, H. (2019).
\newblock Data-driven methods for predictive maintenance of industrial equipment: A survey.
\newblock {\em IEEE Systems Journal}, 13:2213--2227.

\bibitem[Zhou, 2012]{zhou2012ensemble}
Zhou, Z.-H. (2012).
\newblock {\em Ensemble methods: foundations and algorithms}.
\newblock CRC press.

\bibitem[Zio, 2013]{Zio2013PrognosticsAH}
Zio, E. (2013).
\newblock Prognostics and health management of industrial equipment.

\end{thebibliography}

\end{document}